%% file: main.tex
\pdfoutput=1
\documentclass[10pt]{article}
\usepackage[letterpaper,margin=1in,headsep=14pt]{geometry}
\usepackage[T1]{fontenc}
\usepackage[utf8]{inputenc}
\usepackage{amsmath,amsthm,mathtools}
\usepackage[helvratio=0.92]{newtxtext}
\usepackage{newtxmath}
\usepackage{microtype}
\usepackage{booktabs,multirow,array}
\usepackage{graphicx}
\usepackage[table]{xcolor}
\usepackage[authoryear,round]{natbib}
\usepackage{enumitem}
\usepackage{placeins}
\usepackage{longtable}
\definecolor{ptsnavy}{HTML}{1F3A68}
\definecolor{ptsaccent}{HTML}{D9472B}
\definecolor{ptsgray}{HTML}{F3F4F6}
\definecolor{ptsrow}{HTML}{FDEEE8}
\definecolor{ptslink}{HTML}{24539A}
\usepackage{titlesec}
\titleformat{\section}{\sffamily\bfseries\large\color{ptsnavy}}{\thesection}{0.6em}{}
\titleformat{\subsection}{\sffamily\bfseries\normalsize\color{ptsnavy}}{\thesubsection}{0.6em}{}
\titleformat{\paragraph}[runin]{\sffamily\bfseries\small}{}{}{}[.]
\titlespacing*{\section}{0pt}{10pt plus 3pt minus 2pt}{5pt plus 1pt}
\titlespacing*{\subsection}{0pt}{8pt plus 2pt minus 2pt}{3pt plus 1pt}
\titlespacing*{\paragraph}{0pt}{5pt plus 1pt minus 1pt}{0.6em}
\usepackage{caption}
\DeclareCaptionLabelSeparator{bar}{\enspace\textcolor{gray}{|}\enspace}
\usepackage{fancyhdr}
\fancypagestyle{first}{\fancyhf{}\fancyfoot[C]{\sffamily\footnotesize\thepage}}
\setlist{leftmargin=*,itemsep=1.5pt,topsep=2pt}
\usepackage[colorlinks=true,linkcolor=ptslink,citecolor=ptslink,urlcolor=ptslink]{hyperref}
\usepackage[capitalise,noabbrev]{cleveref}
\input{macros}

\newcommand{\ours}{\rowcolor{ptsrow}}

\begin{document}
\thispagestyle{first}
\begin{center}
{\LARGE\bfseries BlinkMimic: Training a Humanoid Motion Tracker\\[2pt] in Minutes on a Single GPU\par}
\vspace{14pt}
{\large Fuyi Chen\par}
\end{center}
\vspace{4pt}
\begin{center}{\large\bfseries Abstract}\end{center}
\vspace{-4pt}
\begin{quote}
\input{sections/abstract}
\end{quote}
\vspace{2pt}
\input{sections/intro}
\input{sections/related}
\input{sections/setup}
\input{sections/method}
\input{sections/experiments}
\input{sections/discussion}
\FloatBarrier
\bibliographystyle{plainnat}
\bibliography{refs}
\clearpage
\appendix
\crefalias{section}{appendix}\crefalias{subsection}{subappendix}
\input{appendix/proofs}
\input{appendix/details}
\input{appendix/tests}

\end{document}

%% file: macros.tex
\theoremstyle{remark}

\theoremstyle{plain}
\newcommand{\E}{\mathbb{E}}

\newcommand{\Nfit}{N_{\mathrm{fit}}}
\newcommand{\Ntail}{N_{\mathrm{tail}}}
\newcommand{\Bnoise}{B_{\mathrm{noise}}}
\newcommand{\polar}{\operatorname{polar}}
\newcommand{\bm}{BlinkMimic}

%% file: sections/abstract.tex
Whole-body motion tracking lets a humanoid robot follow reference motions, but training a tracking policy for a library of motions has usually taken hours of GPU time. SuperMimic showed that the policy can instead be trained by backpropagating through a world model learned from parallel simulation, and InstantTrack added Muon updates, a residual policy around the reference and an averaged deployed policy. We build on both and present \bm{}, a recipe that makes this training faster. On a simulated Unitree G1, \bm{} trains one policy for all 40 LAFAN1 clips in 3.4 minutes on a single RTX 4090, measured until three consecutive checkpoints reach a normalized return of at least 0.82 with no failed clip (mean of three seeds). We split training time into the cost of an iteration, the iterations needed to fit the motions, and the iterations needed to remove the last failures, and assign each component, inherited or new, to one of these terms. The new components are a step size that decreases once training has progressed, resets directed by failures of the deployed policy under deployment conditions, and GPU graph capture of the whole iteration. A leave-one-out ablation with three seeds per component shows which term each component affects, and further measurements of the gradient, the update rule and the failure cases explain these effects. Code and per-seed results are available at \url{https://github.com/fuyichen0725/blinkmimic}.

%% file: sections/intro.tex
\section{Introduction}
\label{sec:intro}

\begin{figure}[t]
\centering
\includegraphics[width=\linewidth]{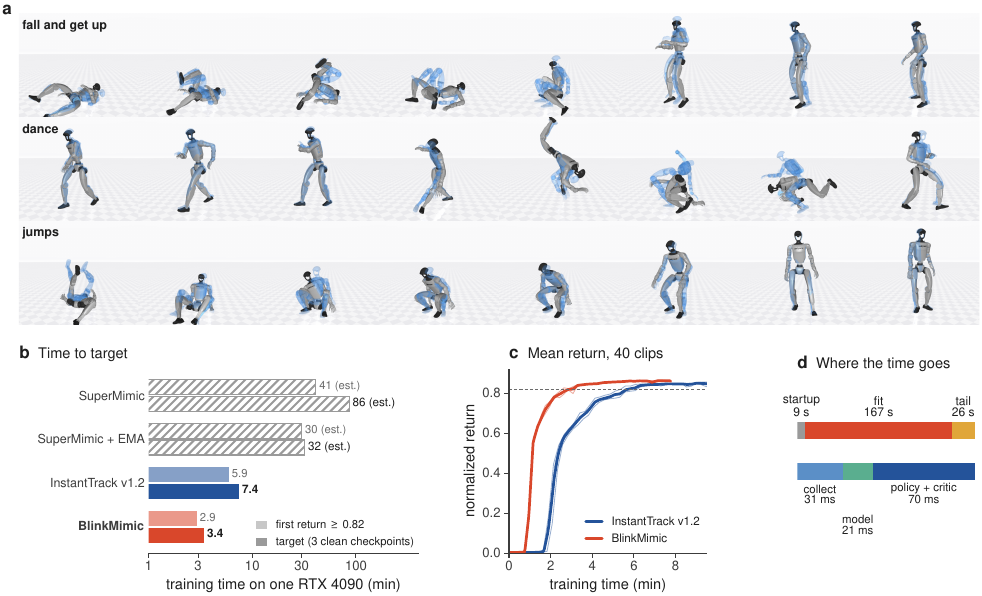}
\caption{\textbf{One policy for 40 LAFAN1 clips in 3.4 minutes on one RTX 4090.} \textbf{a},~The deployed policy of seed 0 at the first checkpoint that meets the target (iteration 1500, 3.32 minutes of training), simulated under evaluation settings; each row shows the 3.6 most dynamic seconds of one clip. Solid: robot. Translucent: reference. \textbf{b},~Training time to the first mean return of 0.82 (light) and to the target of three consecutive clean checkpoints (dark), on the same GPU with the same evaluator. InstantTrack and \bm{}: measured, mean of three seeds. SuperMimic (hatched): estimated as the iterations of InstantTrack's SuperMimic runs times SuperMimic's time per iteration measured on our GPU. \textbf{c},~Mean return over 40 clips $\times$ 4 replicas against training time; thin lines are seeds. \textbf{d},~Where the 3.4 minutes go (mean of three seeds), and the cost of one iteration by stage, measured before the collection step was captured as a graph.}
\label{fig:teaser}
\end{figure}

A humanoid that can follow arbitrary reference motions is a building block for teleoperation, imitation and whole-body skills \citep{peng2018deepmimic,he2024omnih2o,liao2025beyondmimic}. Training such a tracker has been expensive: with on-policy reinforcement learning, a policy for a library of motions takes hours to days on several GPUs \citep{luo2023phc,chen2025gmt}. SuperMimic \citep{liao2026supermimic} took a different route. It learns a differentiable world model of the robot from parallel simulation and improves the policy by backpropagation through time (BPTT) in that model, replacing the high-variance policy-gradient estimate with a pathwise one. This cut the cost of training a tracker substantially and made training time a quantity worth optimizing in its own right. InstantTrack \citep{dao2026instanttrack} then showed in an open implementation that a few changes, an averaged deployed policy, Muon updates \citep{jordan2024muon} and a policy that outputs a correction to the reference, bring training on the 40 LAFAN1 clips \citep{harvey2020lafan} down to minutes. Training in minutes changes how such controllers are developed: a change to the reward, the retargeting or the robot model can be tested within the same hour on one workstation.

This paper builds on both and asks how far, and why, this training can be made faster. Our recipe, \bm{}, trains one policy for all 40 clips to a strict target, three consecutive checkpoints with a mean return of at least 0.82 and no failed clip, in 3.4 minutes on one RTX 4090 (\cref{fig:teaser}). Recipes like this tend to accumulate changes without a clear account of which ones matter. We organize ours around a simple identity: the time to the target is the cost of an iteration times the number of iterations, and the iterations split into a \emph{fit phase}, until the mean return reaches 0.82, and a \emph{tail phase}, until the last failing clips stop failing. The three terms are limited by different things. We assign each ingredient, inherited from SuperMimic and InstantTrack or new, to one of them, and add new ingredients where a term was not yet addressed.

\paragraph{The fit phase is limited by the update rule}
The imagined gradient is accurate over the horizon on which the policy acts, and regression onto a perfect teacher needs no fewer updates than learning from the imagined gradient. What remains is how much each update achieves. Muon updates, here applied to both the policy and the world model, and the residual policy shorten this phase, and we add a step size that starts high and falls once training has progressed.

\paragraph{The tail phase is limited by where the simulator is spent}
The last failures occur at a few states that the deployed policy reaches only after minutes of tracking. These states are learnable: the hardest clip trained alone stops failing early, but in the mix the states before its hard spot receive about 1\% of the practice, and failure statistics taken during training point at the wrong places. A periodic probe with the deployed policy under deployment conditions finds these states directly.

\paragraph{The iteration is limited by the host}
At these model sizes the GPU often waits for the CPU. We capture the whole iteration, including the MuJoCo-Warp \citep{mjwarp2025} physics step and episode resets, as CUDA graphs, and halve the imagination batch, which costs fewer extra iterations than it saves per iteration.

\medskip
A leave-one-out ablation of the final recipe, with three seeds per ingredient, checks that each ingredient acts on the term it was chosen for. Our contributions are:
\begin{itemize}
  \item \textbf{\bm{}}, a single-GPU recipe that trains one policy for 40 LAFAN1 clips in 3.4 minutes, with code and per-seed results (\cref{tab:main}).
  \item \textbf{A design organized by where training time goes}, with a short analysis for each ingredient: how model error enters the gradient, why the reward favors a residual policy, why the step size should fall, why normalized updates produce a batch knee, and why failure statistics from training misdirect practice (\cref{sec:method}).
  \item \textbf{A leave-one-out ablation} on a common baseline (\cref{tab:ablation}) and \textbf{measurements that explain it}: gradient comparisons between the model and the simulator, a perfect-teacher reference, batch forks with a prediction registered before the run, and attribution tests for the tail (\cref{sec:exp}).
\end{itemize}
\Cref{sec:setup} describes the task and the training loop, \cref{sec:method} presents the ingredients term by term, and \cref{sec:exp} reports the result and the ablation and then examines the fit phase and the tail phase.

%% file: sections/related.tex
\section{Related Work}
\label{sec:related}

\paragraph{Humanoid motion tracking}
Physics-based tracking grew out of DeepMimic \citep{peng2018deepmimic} and was scaled to large libraries by PHC \citep{luo2023phc}, MaskedMimic \citep{tessler2024maskedmimic}, OmniH2O \citep{he2024omnih2o}, GMT \citep{chen2025gmt} and BeyondMimic \citep{liao2025beyondmimic}. These works use on-policy RL with massively parallel simulation. SuperMimic \citep{liao2026supermimic} replaced the policy-gradient estimator with BPTT through a learned world model, and InstantTrack \citep{dao2026instanttrack} introduced EMA deployment, Muon and a reference residual in an open implementation. We build on this line.

\paragraph{Differentiable simulation and model-based gradients}
Pathwise gradients through simulators \citep{xu2022shac,georgiev2024ahac} or learned models \citep{deisenroth2011pilco,hafner2020dreamer,hafner2023dreamerv3} trade variance for bias, which can be large at contacts and in chaotic rollouts \citep{suh2022differentiable,metz2021gradients}; model error compounds over the horizon \citep{janner2019mbpo,buckman2018steve}. Mixed estimators interpolate between the two gradient types \citep{parmas2018pipps,suh2022differentiable}.

\paragraph{Optimizers, batch size and step size}
Muon \citep{jordan2024muon,liu2025muon} orthogonalizes the update of each weight matrix; SOAP \citep{vyas2024soap} preconditions in a Kronecker eigenbasis. The noise scale predicts a critical batch for SGD-like training \citep{mccandlish2018empirical}, and its interaction with momentum and normalization is subtle \citep{zhang2019nqm,malladi2022sde,cutkosky2020momentum}. Adaptive methods sit near an edge of stability \citep{cohen2022adaptive}; range tests \citep{smith2017cyclical} and warmup--stable--decay schedules \citep{hu2024minicpm} set the step size in practice.

\paragraph{Allocation and evaluation}
Prioritized sweeping \citep{moore1993prioritized} and level replay \citep{jiang2021plr} spend computation where learning progresses or fails, and imitation learning requires practice on the states the learner itself reaches \citep{ross2011dagger,laskey2017dart}. Time-to-target on stated hardware is the standard of DAWNBench \citep{coleman2017dawnbench} and MLPerf \citep{mattson2020mlperf}; RL comparisons need multiple seeds \citep{agarwal2021rliable}.

%% file: sections/setup.tex
\section{Problem Setup}
\label{sec:setup}

\paragraph{Task}
A Unitree G1 with 29 actuated joints tracks reference clips in MuJoCo-Warp \citep{mjwarp2025} at 50\,Hz. We use the 40 LAFAN1 clips retargeted to the G1 (441{,}140 frames, 2.45 hours; the average clip lasts 3.7 minutes). The policy outputs a normalized action that is scaled and passed as a joint position target to PD controllers. The reward is a sum of Gaussian tracking kernels with an action-rate penalty, and an episode fails when the height or tilt error of the anchor body exceeds a threshold, following SuperMimic \citep{liao2026supermimic}.

\paragraph{Evaluation}
The deployed policy runs deterministically, without domain randomization or noise, from the first frame to the end of every clip, as in SuperMimic and InstantTrack. The normalized return $\bar R\in[0,1]$ accumulates until the first failure. A checkpoint meets the target when $\bar R\ge0.82$ with zero failed clips; \cref{sec:exp} requires this at three consecutive checkpoints.

\begin{figure}[t]
\centering
\includegraphics[width=\linewidth]{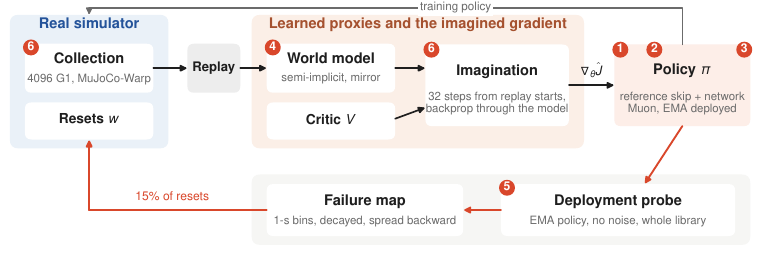}
\caption{\textbf{One iteration of \bm{}.} The real simulator (blue) feeds the learned proxies (orange), and the policy improves through the imagined gradient. Numbers mark the ingredients, inherited or new, in the order of \cref{sec:method}: (1) Muon updates for the policy and the world model; (2) a step size that falls on progress; (3) the reference residual; (4) the semi-implicit, mirror-augmented world model; (5) the deployment-condition probe and reset allocation; (6) the imagination batch and the whole iteration captured as GPU graphs.}
\label{fig:overview}
\end{figure}

\paragraph{Training loop}
\Cref{fig:overview} shows one iteration. In \emph{collection}, 4096 simulated robots step the noisy training policy under domain randomization and store transitions in a replay buffer; finished episodes restart at a frame drawn from a \emph{reset distribution} $w$ over 1-second bins of the library. A learned \emph{world model} $\hat F_\phi$ is fit to short replay windows. In \emph{imagination}, $B$ start states from the buffer are rolled out through $\hat F_\phi$ for $H=32$ steps, and the policy follows the gradient, obtained by BPTT, of
\begin{equation}
\hat J(\theta)=\E_{z_0}\Big[\textstyle\sum_{t<H}\gamma^t\hat\alpha_t\,\hat r_t+\gamma^H\hat\alpha_H V_\psi(\hat z_H)\Big],
\label{eq:proxy}
\end{equation}
where $\hat\alpha_t$ is a product of soft survival probabilities and $V_\psi$ is a critic fit to $\lambda$-returns of the imagined rollouts. As in InstantTrack, the deployed policy is an exponential moving average (EMA, $\beta=0.995$) of $\pi_\theta$ \citep{polyak1992}.

Three distributions over the library should be kept apart: the reset distribution $w$, which we control; the visitation $d$ of the training robots; and the imagination-start distribution $u$, which follows $d$ with the delay of the replay buffer. Only $u$ decides what the policy practises.

%% file: sections/method.tex
\section{BlinkMimic}
\label{sec:method}

Let $c_i$ be the wall-clock cost of iteration $i$ and $N$ the first iteration that meets the target. The training time is
\begin{equation}
T=\sum_{i<N}c_i=\bar c\,(\Nfit+\Ntail),
\label{eq:decomp}
\end{equation}
where the \emph{fit phase} $\Nfit$ ends when the mean evaluation return first reaches 0.82 and the \emph{tail phase} $\Ntail=N-\Nfit$ lasts until the last failing clips stop failing. The three terms respond to different changes, so we design for them separately. \Cref{sec:muon,sec:lr,sec:skip,sec:wm} shorten the fit phase, \cref{sec:alloc} shortens the tail, and \cref{sec:cost} makes each iteration cheaper; \cref{fig:overview} marks where each ingredient sits in the loop. The order follows the ablation in \cref{tab:ablation}. The full configuration is in \cref{app:config} and the derivations behind the short analyses are in \cref{app:proofs}.

\subsection{Muon for the policy and the world model}
\label{sec:muon}
InstantTrack \citep{dao2026instanttrack} introduced Muon \citep{jordan2024muon} to this setting. Muon replaces the update of each hidden weight matrix with $\eta\,UV^\top$, where $U\Sigma V^\top$ is the singular value decomposition of the momentum matrix. The step moves every singular direction by the same amount; it is steepest descent under a limit on the spectral norm of the step. Input, output and bias parameters keep Adam, and the learning rate is matched to Adam's update magnitude \citep{liu2025muon}.

Muon matters here because, as \cref{sec:exp-why} shows, the fit phase is limited by how much each update achieves, not by the quality of the imagined gradient: regressing the same network onto a perfect teacher takes 2.3$\times$ fewer updates with Muon than with Adam. As in InstantTrack, the world model is also trained with Muon, which halves its loss early in training.

\subsection{A step size that starts high and falls on progress}
\label{sec:lr}
A larger step removes the initial error faster but leaves more noise at the end. On a quadratic $L=\tfrac12hx^2$ with noisy gradients of variance $\sigma^2$, the error contracts faster as the step $\eta$ grows (up to $\eta=1/h$), while the remaining excess loss is $\eta\sigma^2/(2(2-\eta h))$, which grows with $\eta$ (\cref{app:floor}). Averaging the iterates, as our EMA policy does, lowers this floor mainly along stiff directions. In world-model training the objective also moves between iterations, which adds to the noise.

We therefore start at the largest policy step that a short screen admits, four times the base rate, and reduce it linearly to the base rate over 300 iterations once the training return first exceeds 0.75, or at iteration 1400 at the latest. The trigger adapts to how fast a seed learns, and the deadline covers seeds whose return rises late.

\subsection{A residual policy}
\label{sec:skip}
Following InstantTrack's reference residual, the policy outputs a correction to the next reference joint target, $a=a^{\mathrm{ref}}+N_\theta$, with the last layer initialized near zero; the observation and the MDP are unchanged. Here we explain why this helps, and the reason lies in the reward. A tracking kernel $r=\exp(-e^2/\sigma^2)$ has the slope $|\partial r/\partial e|=(2/\sigma)\sqrt x\,e^{-x}$ with $x=e^2/\sigma^2$. The slope peaks at $x=\tfrac12$, falls to 30\% of the peak at $x=2.5$ and to 3.5\% at $x=5$. A policy far from the reference therefore gets almost no gradient from its largest errors, while a residual policy starts where every kernel is informative (\cref{fig:grad}c).

\subsection{A world model with the simulator's structure}
\label{sec:wm}
The policy gradient flows through the model's Jacobians, so what matters is how well these Jacobians match the simulator along imagined rollouts. Writing $x_t$ for the sensitivity of the state to the policy parameters, the error of the model's sensitivity is a sum of the Jacobian errors at each earlier step, each propagated forward by the model's own dynamics (\cref{app:err}):
\[
\hat x_t-x_t=\sum_{k<t}\hat\Phi(t,k{+}1)\big[\Delta A_k\,x_k+\Delta B_k\,\Pi_k+\hat B_k\,\Delta\Pi_k\big].
\]
Here $\Delta A_k$ and $\Delta B_k$ are the model's errors in the closed-loop state Jacobian and the action Jacobian, $\Pi_k=\partial\pi/\partial\theta$, $\Delta\Pi_k$ is the change of $\Pi_k$ caused by the model's different state, and $\hat\Phi(t,k)$ is the product of the model's closed-loop Jacobians from step $k$ to $t$.
Two consequences guide the design. Jacobian errors matter only where imagination goes and where later dynamics amplify them, so the model should be accurate on the policy's own states. And a small one-step prediction error does not imply a small gradient error, so structure that shapes the Jacobians is worth more than extra accuracy of the prediction itself.

The model therefore uses the integrator's structure, as does the implementation released with SuperMimic. It predicts a velocity change and integrates position semi-implicitly, plus a small learned position correction; if it shared the integration step exactly, its position Jacobian errors would be a factor $\Delta t$ smaller than its velocity errors. Half of every training window is replaced by its left--right mirror image, a symmetry of the robot that costs no simulation. In the final recipe, where the model is also trained with Muon, removing the structure and the mirror no longer lengthens the fit phase but lowers the return at which the policy settles, and with it the margin above the target (\cref{sec:exp-ablation}).

\subsection{Spending the simulator on deployment failures}
\label{sec:alloc}
After the mean return reaches the target, a few clips still fail at a few places. Let $p_i$ be the probability that the \emph{deployed} policy fails in 1-second bin $i$ under \emph{deployment} conditions, given that it got there. When failures are rare, the number of failed clips is about $\sum_ip_i$. If practice on a bin lowers its $p_i$, the fastest way to remove failures is to practise the bins with the highest $p_i$ first (\cref{app:alloc}). Transfer between clips makes this only a guide, but it says what should drive the reset distribution: failure rates of the deployed policy under deployment conditions.

Failure statistics collected during training measure something else. The training policy is noisy, randomized and not averaged. If resets are weighted by those statistics, bins that cannot be learned under randomization keep absorbing resets, while bins that fail only under deployment conditions receive no more than the uniform share (\cref{app:mismatch}). \bm{} therefore measures deployment failures directly. Every 125 iterations from iteration 1000, a separate simulator sweeps the whole library with the EMA policy under deployment conditions, using one world per 300-frame tile started on the reference 200--450 frames before the tile. The first failure of each world is counted in its bin, counts decay by 0.7 per sweep, and each count is spread over the failing bin and the 10 bins before it, because a failure develops from the policy's own drift. The resulting map receives 15\% of all resets; the rest keep the sampler's uniform, low-return and training-failure terms (\cref{app:config}). The probe needs no prior knowledge of which clips are hard.

\subsection{Making each iteration cheap}
\label{sec:cost}
\paragraph{Imagination batch}
Because Muon fixes the length of the step, a smaller batch does not make the step noisier in size; noise only rotates it away from the signal. A signal direction is lost once its strength falls below a detection threshold set by the noise, as known for spiked random matrices \citep{benaych2012svd}. The number of steps to a target then stays nearly flat as the batch shrinks and rises sharply once the step's overlap with the signal nears a curvature cost (\cref{app:knee}), a \emph{knee} rather than the smooth $1+\Bnoise/B$ curve of SGD \citep{mccandlish2018empirical}. Since the position of the knee depends on the model, we measure it with short forks and take $B=2048$, the smallest batch that leaves the tail intact (\cref{sec:exp-why}).

\paragraph{Removing the host from the loop}
At batch 2048, the updates and collection take 122\,ms on an RTX 4090, but the MuJoCo-Warp kernels of a collection step add up to only about 11\,ms of its 31\,ms; the rest is launch gaps and host synchronizations. InstantTrack already captures the two update phases as CUDA graphs; we capture them from the first iteration and add the collection step, with the policy inference, physics step and reward in one CUDA graph with episode resets done by fixed-shape operations inside that graph. The probe sweep is also a graph and runs alongside training over 20 iterations. None of this changes the learning algorithm: after a masked reset every simulator field matches the original bit for bit, except joint velocities, which differ by at most $2.4\times10^{-7}$ (\cref{app:systems}). In two alternating paired runs on the same GPU, without checkpoints or probes, the collection graph lowers the steady-state iteration from 127.5 to 118.5\,ms ($-$7\%) and shortens the first 700 iterations by 18\,s.

%% file: sections/experiments.tex
\section{Experiments}
\label{sec:exp}

\paragraph{Protocol}
Every checkpoint is evaluated on all 40 clips from the first frame, with four replicas per clip, because the simulator is not bitwise reproducible and a clip near its margin can flip between evaluations. A checkpoint is \emph{clean} when the mean return is at least 0.82 and all 160 rollouts survive. The target is three consecutive clean checkpoints, reported at the first of them, as in InstantTrack \citep{dao2026instanttrack}; we also report the first checkpoint with return at least 0.82, which marks the end of the fit phase. We checkpoint every 100 iterations (about 13\,s). Training time includes compilation, graph capture, checkpoints and probes and excludes the offline evaluation. Runs that report time have the RTX 4090 to themselves, and baselines are re-run on the same GPU with the same evaluator and checkpoint interval. The ablation runs on a different GPU type and compares iterations, each row against the full recipe run on that same GPU type. Because the simulator is not bitwise reproducible across GPUs, the full recipe needs more iterations there than on the RTX 4090 (1422 against 1306 to 0.82, and 2067 against 1533 for the three-checkpoint rule), so numbers are compared only within a table. Seed counts are stated for every number; several diagnostics use one seed and are marked as such.

\subsection{Main result}
\label{sec:exp-main}
\input{tables/main}

\bm{} reaches the target in 3.32, 3.28 and 3.53 minutes for seeds 0, 1 and 2 (\cref{tab:main}). The first checkpoint at return 0.82 comes at 2.94 minutes on average, so the tail phase costs 26 seconds. One iteration costs 135\,ms including probes and checkpoints. At the target, the RMS position error of the 14 tracked bodies is 4.1\,cm and the mean joint-angle error 4.0\textdegree{}; by iteration 3500 the body error falls to 3.4\,cm (\cref{tab:accuracy}). Meeting the target does not guarantee that later checkpoints stay clean: in seed 1, \texttt{fallAndGetUp2\_subject2} fails in all four replicas at iteration 3100, long after the step size has settled. We return to such late flips in \cref{sec:discussion}; \cref{app:lineage} lists the earlier recipes.

\subsection{What each ingredient contributes}
\label{sec:exp-ablation}
\input{tables/ablation}

We remove one ingredient at a time from the release recipe and keep everything else (\cref{tab:ablation}). Outside their own rows, the step-size multiple and the batch stay at the values tuned with all ingredients present. Each row has three seeds, and all 24 runs share one GPU type.

\paragraph{Fit phase}
Muon on the policy is the one ingredient the recipe cannot do without. With Adam in its place, the return reaches 0.80 by iteration 3000 and stays at 0.81 up to iteration 6000 in all three seeds, with no failed rollout at the end: the policy survives but does not track precisely enough. The row keeps the $\times4$ step and the batch of 2048, so it measures what Muon contributes inside the recipe, not the best Adam can do with its own settings. Removing Muon on the world model, the progress-triggered step size or the reference residual lengthens the fit phase by 24\%, 14\% and 23\%, and every seed is slower than every seed of the full recipe. The three-checkpoint rule moves in the same direction but is noisier ($+$32\%, $+$15\%, $+$8\%), because it also depends on the tail. Without these ingredients the policy also settles at a lower return (0.843--0.856 against 0.860).

\paragraph{Return level}
Removing the semi-implicit structure and the mirror leaves the fit phase unchanged (1428 against 1422 iterations) but lowers the mean return over iterations 2000--3500 in every seed (0.8535--0.8548 against 0.8581--0.8628), and two of three seeds fail more often late. In the final recipe the structure therefore acts on the margin above the target. In earlier configurations without Muon on the world model it shortened the fit phase by 19\% (\cref{tab:lineage}). The row removes structure and mirror together, so their shares are not separated.

\paragraph{Tail phase}
Without the probe, late failures double, from 9 to 18 rollouts over three seeds (1 / 8 / 0 against 10 / 0 / 8). This is the direction found in an earlier recipe (\cref{sec:exp-tail}), but the counts are small and the seeds overlap, so three seeds do not establish the effect here, and the three-checkpoint rule is not slower without the probe. The probe also costs time: on the RTX 4090 its earlier synchronous form added 20\,ms to 121\,ms per iteration (\cref{tab:earlier}).

\paragraph{Iteration cost}
Restoring the batch of 4096 reduces the iterations to 0.82 by 20\% but cost 41\% more per iteration on the RTX 4090 in an earlier configuration (172 against 122\,ms, \cref{tab:lineage}), so the time to target grows. The paired comparison on the RTX 4090 in \cref{sec:exp-why} gives the same answer.

The rest of this section asks why the ingredients act where they do: what limits the fit phase (\cref{sec:exp-why}) and where the tail comes from (\cref{sec:exp-tail}).

\subsection{Why the fit phase is limited by the update rule}
\label{sec:exp-why}\label{sec:exp-grad}\label{sec:exp-fit}
The ablation shows that the update rule (Muon, step size) and the residual policy set the length of the fit phase. This section follows one line of argument: the imagined gradient is accurate where it matters, a perfect teacher is not faster, so what remains is how much each update achieves.
\begin{figure}[t]
\centering
\includegraphics[width=\linewidth]{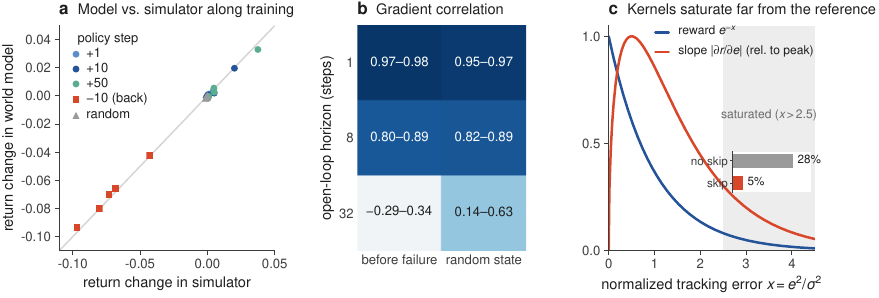}
\caption{\textbf{The imagined gradient.} \textbf{a},~Change of the 32-step return, in the simulator and in the world model, for policies a few iterations ahead of the current one, for a step backward along the path, and for random directions of equal norm (2048 on-policy states, 5 checkpoints from iteration 1000 to 3000, one seed). \textbf{b},~Correlation between the model's directional derivatives of the open-loop return and central differences in the simulator (48 states; ranges over checkpoints). \textbf{c},~A Gaussian tracking kernel and its slope. Inset: share of body-orientation errors past $x=2.5$ at iteration 1000 under training conditions, with and without the reference residual (one seed).}
\label{fig:grad}
\end{figure}

\paragraph{The imagined gradient is accurate where the policy acts}
We test the model against the simulator in two ways. Along the actual training path (\cref{fig:grad}a), the change of return predicted by the model matches the simulator, and at no checkpoint is the model more optimistic about the next policy than about the current one (model return minus simulator return 0.0151 against 0.0157 at iteration 1000, about 0.6\% of the return). Online training thus leaves no measurable model exploitation at the scale of one update; it does appear when the model is frozen, where the return stops near 0.70 with 14 clips failing (one seed). A step \emph{backward} along the path loses 0.04--0.10, while steps forward change the return by little after iteration 1000. Random directions of equal norm change nothing. The optimizer moves along stiff directions and makes its net progress over hundreds of iterations, as expected near an edge of stability \citep{cohen2022adaptive}. The EMA gains a further 0.012 over the raw policy.

\Cref{fig:grad}b shows the limit of the model. The action gradient over 1 and 8 steps correlates at 0.95--0.98 and 0.80--0.89 with the simulator, also just before failures, but over 32 steps the correlation is low. As the sensitivity analysis in \cref{sec:wm} suggests, the error compounds with the horizon. The critic carries much of the signal, and its slope is unreliable (correlation with the true change of return at most 0.11), yet removing it costs 50\% more iterations. The gradient is thus accurate at the horizon where the policy can act, and the long horizon mostly delivers the value estimate.

\begin{figure}[t]
\centering
\includegraphics[width=\linewidth]{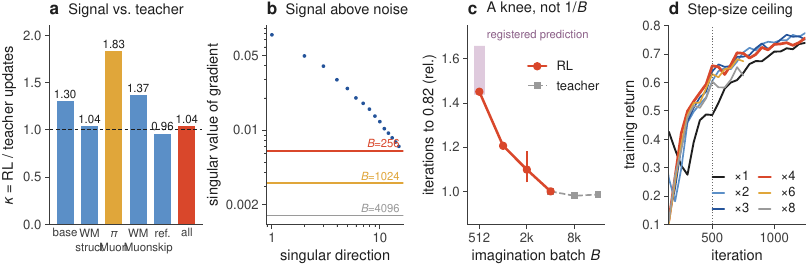}
\caption{\textbf{The update rule limits the fit phase.} \textbf{a},~$\kappa$, RL updates to reach 0.82 divided by updates of a regression onto a perfect teacher with the same network and optimizer. Blue: changes to the signal; amber: an update-rule change alone, which helps the teacher more than RL. Run lengths and seeds are in \cref{app:lineage}. \textbf{b},~Singular values of one policy gradient matrix (iterations 700--1600, one seed) and the detection threshold for three batch sizes. \textbf{c},~Iterations to 0.82 against batch, relative to 4096, for RL (2048: three seeds, bar: range; 1024, 512: one seed) and for the teacher; shaded: range registered before the 512 run. \textbf{d},~Training return for step-size multiples ($\times$1: three-seed mean; others one seed).}
\label{fig:fit}
\end{figure}

\paragraph{A perfect teacher is not faster}
To ask how much better the signal could be, we train the same network with the same optimizer by regression onto the actions of a converged policy, with DAgger-style data collection, and define $\kappa$ as the ratio of RL updates to teacher updates needed to reach 0.82. The teacher is a reference, not a bound. \Cref{fig:fit}a traces $\kappa$ along the history of the recipe. The first configuration needs 30\% more updates than regression; the structured world model brings $\kappa$ to 1.04. Muon on the policy speeds up the teacher by 2.3$\times$ (3494 to 1478 updates, and SOAP matches it at 1505) but RL by less, so $\kappa$ rises to 1.83. Muon on the world model and the reference residual then bring it back to about 1.0. Before the step-size schedule, the recipe needed 1549 iterations at batch 4096, against 1478--1499 for the teacher; with the schedule, the release recipe needs 1306 at batch 2048, fewer than the teacher. A better signal therefore has little left to give, while a better update rule, as the teacher's 2.3$\times$ gain from Muon shows, can still help.

\paragraph{The batch knee}
In-batch noise is small. Cosines between the gradients of two disjoint halves of the batch lie mostly in 0.62--0.87 under Muon, and the teacher gains only 3\% from a batch of $\times2$ or $\times4$. Most singular directions of the gradient that carry energy remain above the detection threshold at smaller batches (\cref{fig:fit}b). A measurement of the resolved signal \emph{count} alone did not explain the knee, however, so we fit the single parameter $\rho=0.53$ of the knee model (\cref{app:knee}) to the fork at $B=1024$. Before the run at $B=512$ we registered a prediction of 2280--2630 iterations; the run took 2301 (\cref{fig:fit}c, one seed). A further fork that kept 4096 samples for the critic and used 1024 for the policy kept about 84\% of the slowdown (1.17$\times$), so the knee comes mostly from the policy gradient and not from critic data. At $B=2048$ the wall time to 0.82 falls by 17\% with every seed faster, and we stop there because smaller batches increase late failures. In the release recipe, with its larger step, the cost of halving the batch is higher: 1422 against 1138 iterations to 0.82 (\cref{tab:ablation}), a 25\% increase, still smaller than the 29\% saved per iteration (122 against 172\,ms in \cref{tab:lineage}).

\paragraph{Step size}
A step-size screen shows the same trade-off. At iteration 500 the return rises from 0.606 at $\times2$ to 0.659 at $\times4$ and falls to 0.603 at $\times8$, without divergence (\cref{fig:fit}d). By iteration 1250, $\times2$ has caught up, consistent with the higher noise floor of a large step (\cref{sec:lr}). In an earlier recipe, a progress-triggered decay from $\times4$ reduced iterations by 19\% against a constant step (three seeds; 14\% in the final recipe, \cref{tab:ablation}), whereas a cosine schedule with a fixed horizon gave no gain.

\subsection{The tail phase}
\label{sec:exp-tail}
\begin{figure}[t]
\centering
\includegraphics[width=\linewidth]{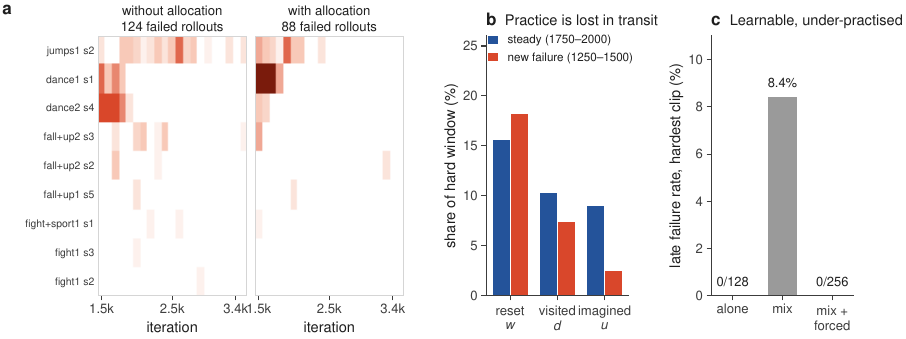}
\caption{\textbf{The tail is a practice problem.} \textbf{a},~Failed rollouts per clip and checkpoint over iterations 1500--3500, summed over three seeds and four replicas (maximum 12 per cell), for recipe B without and with allocation; the tail counts in the text start at 2000. \textbf{b},~Share of a hard window in the reset distribution, the visited states and the imagination starts (one seed). \textbf{c},~Late failure rate of the hardest clip trained alone (iterations 1250--3000), in the 40-clip mix (mean over iterations 2500--3000), and in the mix with 10\% of resets forced onto the window before the hard spot (iterations 1750--3000). One seed each.}
\label{fig:tail}
\end{figure}

\paragraph{Where the tail comes from}
Before allocation, most late failures came from one clip, a floor roll followed by a stand-up in \texttt{jumps1\_subject2}. Repairing reference frames, training 10\% of episodes under deployment conditions, anchoring the policy (drift $-$78 to $-$84\%), and adding real data at the failure points to the world model each left the failures unchanged. The test that separated the hypotheses is shown in \cref{fig:tail}c. Trained alone, the clip fails in 16 of 16 rollouts up to iteration 1000 and in 0 of 128 from iteration 1250 on; in the mix, the 10 seconds before the hard spot receive 0.6--1.5\% of resets. Forcing 10\% of resets onto that window gives 0 of 256 failures at the six checkpoints from 1750 to 3000, while the same seed without the change fails 19\%, 3.9\% and 2.3\% of the time at 2500, 2750 and 3000.

\paragraph{Practice is lost on the way}
An audit of one run traced a hard window from resets to imagination (\cref{fig:tail}b). In steady state its share falls from 15.6\% of resets to 10.3\% of visited states and 9.0\% of imagination starts, mostly because episodes started in the window run past it. When a failure is new, the buffer lags: 18.2\% of resets become only 2.5\% of imagination starts until the buffer turns over, about 400 iterations later.

\paragraph{Allocation in two recipes}
We registered an ablation before running it on recipe B, which had a synchronous probe (\cref{tab:earlier}): allocation is effective if tail failures fall by at least 40\% on average and in at least two of three seeds. Tail failures fall from 23, 22 and 10 to 2, 10 and 2 (\cref{fig:tail}a), so the criterion is met, and the three-checkpoint rule improves from 6.1 to 4.9 minutes. In the release recipe the other ingredients leave fewer late failures to remove: without the probe, 18 rollouts fail over three seeds, against 55 in recipe B without it, and the probe halves them to 9 (\cref{sec:exp-ablation}). The direction is the same, but at these counts three seeds no longer separate the two arms.

\paragraph{What did not help}
\Cref{tab:neg} in the appendix lists the alternatives that did not shorten training, including critic variants, shorter or attenuated horizons, mixed gradient estimators, planning teachers, wider networks and prioritized replay.

%% file: tables/main.tex
\begin{table}[t]
\centering
\caption{\textbf{Time to target on one RTX 4090.} 40 clips $\times$ 4 replicas, a checkpoint every 100 iterations, training clock including compilation and probes; seeds 0 / 1 / 2 with the mean in parentheses. Times are in minutes. \emph{3 clean checkpoints}: the first of three consecutive checkpoints with return $\ge0.82$ and no failed rollout. \emph{Tail failures}: failed rollouts summed over checkpoints 2000--3500.}
\label{tab:main}
\small
\setlength{\tabcolsep}{3.2pt}
\resizebox{\linewidth}{!}{%
\begin{tabular}{@{}l c c c c c c@{}}
\toprule
Configuration & ms/it & Iterations to 0.82 & First $\ge$0.82 & 3 clean checkpoints & Tail failures & Final return \\
\midrule
InstantTrack v1.2 (re-run) & 147 & 2408 / 2741 / 2492 & 5.6 / 6.3 / 5.8 (5.9) & 7.3 / 8.5 / 6.3 (7.4) & 80 / 126 / 113 & 0.856 \\
\ours \textbf{BlinkMimic (release)} & 135 & 1292 / 1456 / 1169 & 2.89 / 3.28 / 2.64 (2.94) & 3.32 / 3.28 / 3.53 (3.38) & 2 / 6 / 0 & 0.861 \\
\bottomrule
\end{tabular}}
\end{table}

%% file: tables/ablation.tex
\begin{table}[t]
\centering
\caption{\textbf{Leave-one-out ablation of the release recipe.} Each row removes one ingredient from the full recipe (the last row restores the earlier batch); three seeds per row, all runs on the same GPU type with the same code and evaluator ($40\times4$ rollouts, a checkpoint every 100 iterations). \emph{It.\ to 0.82}: mean, interpolated, with the change against the full recipe. \emph{3 clean}: first of three consecutive clean checkpoints, per seed. \emph{Tail}: failed rollouts summed over checkpoints 2000--3500, per seed. \emph{Return}: mean over checkpoints 2000--3500. Times on the RTX 4090 are in \cref{tab:main}.}
\label{tab:ablation}
\small
\setlength{\tabcolsep}{3.5pt}
\resizebox{\linewidth}{!}{%
\begin{tabular}{@{}l c c c c c@{}}
\toprule
Removed & Term & It.\ to 0.82 & 3 clean (iteration) & Tail & Return \\
\midrule
\ours \textbf{Full recipe} &  & 1422 & 2100 / 2200 / 1900 & 1 / 8 / 0 & 0.860 \\
\midrule
$-$ Muon on the policy (Adam instead) & $\Nfit$ & not by 6000 & -- / -- / -- & 15 / 3 / 18 & 0.796 \\
$-$ Muon on the world model & $\Nfit$ & 1768 ($+$24\%) & 3000 / 3100 / 2100 ($+$32\%) & 24 / 22 / 7 & 0.843 \\
$-$ Progress-triggered step size (constant $\times1$) & $\Nfit$ & 1616 ($+$14\%) & 3000 / 1800 / 2300 ($+$15\%) & 20 / 3 / 6 & 0.856 \\
$-$ Reference residual & $\Nfit$ & 1751 ($+$23\%) & 1800 / 2700 / 2200 ($+$8\%) & 0 / 20 / 8 & 0.845 \\
$-$ Semi-implicit world model and mirror & $\bar R$ & 1428 ($\pm$0\%) & 1800 / 2900 / 1900 ($+$6\%) & 0 / 15 / 13 & 0.854 \\
$-$ Deployment-condition allocation & $\Ntail$ & 1350 ($-$5\%) & 2000 / 2000 / 1700 ($-$8\%) & 10 / 0 / 8 & 0.861 \\
Imagination batch 4096 instead of 2048 & $\bar c$ & 1138 ($-$20\%) & 1700 / 2000 / 2000 ($-$8\%) & 1 / 0 / 0 & 0.864 \\
\bottomrule
\end{tabular}}
\end{table}

%% file: sections/discussion.tex
\section{Discussion and Limitations}
\label{sec:discussion}

\paragraph{What limits the recipe now}
The fit phase takes more than 80\% of the 3.4 minutes. It already runs at about the perfect-teacher rate, and the ablation shows that every fit-phase ingredient still matters, so further gains there must come from a better update rule, from less for the network to learn from scratch, or from cheaper updates. The clearest open problem is reliability. A checkpoint that is clean at one time can fail later (\cref{sec:exp-main}), and the ablation suggests that the margin above the target, which the world-model structure raises, matters for this alongside the allocation of practice. We regard late flips, not mean speed, as the most valuable target for future work.

\paragraph{Scope of the evidence}
We use one robot and one motion library, report training time on one GPU type, and do not report real-robot deployment. Three seeds cannot resolve differences below about 15\% in the strict rule, so we give every seed. In the final recipe this applies to the allocation: it halves late failures, as it reduced them in an earlier recipe, but the remaining counts are too small for three seeds to separate. The ablation keeps the settings tuned for the full recipe, so a removed ingredient may do better with settings of its own, and the structure and the mirror are removed together. Several diagnostics rest on one seed and were measured on earlier configurations of the recipe, as marked. The asynchronous probe assumes an otherwise idle GPU: when another process shares the GPU, training with it can crash, which the synchronous probe avoids at a cost in speed. The perfect-teacher ratio $\kappa$ is a reference and not a lower bound, since a different update rule changes both numerator and denominator. The analyses in \cref{sec:method} are idealized: they say what to measure and how to read it, and the design choices themselves rest on measurements.

\paragraph{Conclusion}
\bm{} trains one policy for 40 LAFAN1 clips in 3.4 minutes on a single RTX 4090. Splitting the time into iteration cost, fit and tail shows what to work on: the update rule during the fit phase, the allocation of the real simulator during the tail, and the removal of host latency throughout. The split itself does not depend on the robot or the motion library, so the same accounting can be applied to other world-model training setups; whether the same ingredients matter there is a question for measurement, not something our single benchmark settles. We release the recipe and per-seed results to make such comparisons easy.

%% file: appendix/proofs.tex
\section{Derivations}
\label{app:proofs}

\paragraph{Notation}
$z_t$ is the augmented Markov state, $a_t=\pi_\theta(z_t)$, $D_t=F_z$, $B_t=F_a$, $A_t=D_t+B_t\pi_z$, and $x_t=\partial z_t/\partial\theta$. A hat denotes quantities of the learned model along its own rollout.

\subsection{Sensitivity error of the world model}
\label{app:err}
Assume that both rollouts stay where $F$, $\hat F$ and $\pi_\theta$ are differentiable and start from the same state with zero sensitivity. Differentiating gives $x_{t+1}=A_tx_t+B_t\Pi_t$ and $\hat x_{t+1}=\hat A_t\hat x_t+\hat B_t\hat\Pi_t$. With $e_t=\hat x_t-x_t$, $\Delta A_t=\hat A_t-A_t$, $\Delta B_t=\hat B_t-B_t$, $\Delta\Pi_t=\hat\Pi_t-\Pi_t$, subtraction gives $e_{t+1}=\hat A_te_t+\Delta A_tx_t+\Delta B_t\Pi_t+\hat B_t\Delta\Pi_t$; unrolling from $e_0=0$ gives the expression in \cref{sec:wm}, with $\hat\Phi(t,k)=\hat A_{t-1}\cdots\hat A_k$.

The last term cannot be dropped: take $F=a$, $\hat F=a+1$, $\pi_\theta(s)=\theta s$, $s_0=1$, $\theta=1$; then $A=\hat A$ and $B=\hat B$, yet the two-step sensitivities differ by 1. A small prediction error also does not imply a small gradient error: $\hat F=F+\varepsilon\sin(z/\varepsilon^2)$ has function error $\varepsilon$ and derivative error $1/\varepsilon$.

\subsection{Semi-implicit structure}
Suppose the simulator integrates $v_{t+1}=v_t+\Delta t\,f(q_t,v_t,a_t)$, $q_{t+1}=q_t+\Delta t\,v_{t+1}$, and the model shares this step and learns only $\hat f$. The Jacobian errors of $(q,v)$ then have the block form $\big[\begin{smallmatrix}\Delta t^2\,\delta\\ \Delta t\,\delta\end{smallmatrix}\big]$ with $\delta=\partial(\hat f-f)$, so position errors are a factor $\Delta t$ smaller than velocity errors. Our model adds a small learned position correction, so the structure is encouraged, not exact.

\subsection{Muon as steepest descent}
For $G=U\Sigma V^\top$, the minimizer of $\langle G,\Delta\rangle$ subject to $\|\Delta\|_{\mathrm{op}}\le\eta$ is $-\eta UV^\top$ \citep{jordan2024muon}. Muon applies this step to the momentum matrix using an approximate orthogonalization.

\subsection{Batch knee under normalized steps}
\label{app:knee}
Write the Muon step as $\eta\,\polar(M)$ with $\polar(M)=UV^\top$ for $M=U\Sigma V^\top$, and model the momentum as $M=\bar G+E$, with i.i.d.\ noise $E$ whose per-entry variance scales as $1/B$. With a local quadratic model, one step decreases the loss by $\eta\langle\bar G,\polar(M)\rangle-\tfrac{\eta^2}{2}\langle\polar(M),H\polar(M)\rangle$. Since $\polar(M)$ has unit singular values, we treat the curvature term as a constant $C$ independent of $B$; this is the assumption that distinguishes the result from SGD, where noise inflates the step. Random-matrix results for spiked models \citep{benaych2012svd} give the overlaps. With $\tau^2$ the per-entry noise variance of $M$, $\theta_i=s_i/(\tau\sqrt q)$ and $c=p/q$ for a $p\times q$ matrix, $a_i=0$ for $\theta_i^4\le c$, and otherwise
\[
a_i^2=\Big[1-\tfrac{c(1+\theta_i^2)}{\theta_i^2(\theta_i^2+c)}\Big]\Big[1-\tfrac{c+\theta_i^2}{\theta_i^2(\theta_i^2+1)}\Big].
\]
Noise directions contribute zero on average, so $\langle\bar G,\polar(M)\rangle=\sum_is_ia_i(B)=:A(B)$, where $\bar G=\sum_is_iu_iv_i^\top$ and $a_i\in[0,1]$ is the overlap of the step with signal direction $i$. Equal progress with per-step gain $\eta A(B)-C$ requires $S(B)\propto1/(\eta A(B)-C)$. With $f(B)=A(B)/A(B_0)$ and $\rho=C/(\eta A(B_0))$, dividing by the value at $B_0$ gives
\[
\frac{S(B)}{S(B_0)}=\frac{1-\rho}{f(B)-\rho},
\]
which rises sharply once $f(B)$ approaches $\rho$. We measured the singular spectra at $B_0=4096$ and summed over layers. The model covers only directions detectable at $B_0$, assumes $C$ is independent of $B$, and is tested by a single registered prediction on one seed. With $\rho$ fitted to one fork, $S(2048)$ is predicted to be 1.07, against a measured 1.006 for the single fork and 1.10 for the three-seed mean.

\subsection{Saturation of the tracking kernel}
\label{app:sat}
$\partial r/\partial e=-(2e/\sigma^2)\exp(-e^2/\sigma^2)$. Substituting $e=\sigma\sqrt x$ gives the magnitude $(2/\sigma)\sqrt x\,e^{-x}$, maximized at $x=\tfrac12$ with value $(2/\sigma)\,0.429$. At $x=2.5$ the ratio to the peak is $\sqrt{2.5}\,e^{-2.5}/0.429=0.30$; at $x=5$ it is $\sqrt5\,e^{-5}/0.429=0.035$.

\subsection{Noise floor and averaging}
\label{app:floor}
For the noisy step $x_{k+1}=(1-\eta h)x_k+\eta\xi_k$ with $\mathrm{Var}\,\xi=\sigma^2$, let $\phi=1-\eta h$ with $|\phi|<1$. The stationary variance $V$ satisfies $V=\phi^2V+\eta^2\sigma^2$, so $V=\eta\sigma^2/(h(2-\eta h))$ and $\E L=\tfrac12hV=\eta\sigma^2/(2(2-\eta h))$. For the EMA $y_k=(1-\beta)\sum_{j\ge0}\beta^jx_{k-j}$, use $\mathrm{Cov}(x_k,x_{k+l})=V\phi^{|l|}$ to get
\[
\mathrm{Var}\,y=(1-\beta)^2V\sum_{i,j\ge0}\beta^{i+j}\phi^{|i-j|}=V\,\frac{(1-\beta)(1+\beta\phi)}{(1+\beta)(1-\beta\phi)},
\]
which is smallest relative to $V$ when $\phi$ is small, that is, along stiff directions.

\subsection{Equal marginal gains}
\label{app:alloc}
Minimize $\sum_ip_i(n_i)$ subject to $\sum_in_i=R$, $n_i\ge0$, with $p_i$ convex and decreasing. The KKT conditions give $-p_i'(n_i)=\mu$ for $n_i>0$ and $-p_i'(0)\le\mu$ otherwise. For $p_i(n)=p_i^0e^{-n/\tau}$ the marginal gain is $p_i(n)/\tau$, so a greedy rule picks the bin with the highest failure probability. Transfer between clips violates separability, and consecutive hard spots violate convexity.

\subsection{Allocation by mismatched failure statistics}
\label{app:mismatch}
Let the reset weights be $w_i\propto(1-\alpha)\,g(\hat p^M_i)+\alpha/N$ with $g$ increasing and $g(0)=0$, where $\hat p^M_i$ are failure estimates taken under a condition $M$, and let $S$ be bins with failure probability at least $\delta>0$ under $M$ at all times and let the estimator keep $\hat p_i^M\ge\delta'>0$ on $S$. If the remaining bins converge, $\hat p_i^M\to0$ for $i\notin S$, then $g(\hat p_i^M)\to0$ off $S$ while $g(\hat p_i^M)\ge g(\delta')>0$ on $S$. The non-uniform part of $w$ therefore concentrates on $S$, whose share tends to $(1-\alpha)+\alpha|S|/N$. A bin with $p_i^E>0=p_i^M$ has $g(\hat p_i^M)\to0$ and keeps only $\alpha/N$.

%% file: appendix/details.tex
\section{Configuration}
\label{app:config}
\begin{center}\small
\begin{tabular}{@{}lp{13.2cm}@{}}
\toprule
Simulator & MuJoCo-Warp via mjlab, 50\,Hz control, 4096 robots \\
Replay & 1000 steps per robot ($\approx$400 iterations); update-to-data 0.4 \\
World model & MLP $3\times1024$, semi-implicit, mirror 0.5, 5 batches of windows per iteration, Muon \\
Imagination & $B=2048$ starts, $H=32$, $\gamma=0.99$, $\lambda=0.95$, soft survival \\
Critic & 5 epochs per iteration on imagined $\lambda$-returns \\
Policy & encoder/decoder 2048-1024-512-256, history 10, 10 goal frames, reference residual, Muon (lr $4\times10^{-4}$, Adam-matched), grad-norm clip 1 \\
Step size & policy $\times4$ until training return $\ge0.75$ or iteration 1400, then linear to $\times1$ over 300 iterations \\
Deployment & EMA 0.995 \\
Resets & uniform, low-return softmax and training-failure terms (failure reset 0.3), plus 15\% deployment-failure map \\
Probe & every 125 iterations from 1000; 1489 worlds (one per 300 frames), lead 200--450 frames, 1-s bins, decay 0.7, spread 10 bins; asynchronous over 20 iterations \\
Systems & update graphs from iteration 1; collection step and resets as one CUDA graph; 3500 iterations, checkpoint every 100 \\
\bottomrule
\end{tabular}
\end{center}

\section{Systems Details}
\label{app:systems}
\paragraph{Time in an iteration}
Before the collection graph, an iteration at batch 2048 took 122\,ms: policy and critic updates 70\,ms, world-model updates 21\,ms, and collection 31\,ms. The GPU was fully idle for only 8--11\% of the iteration, so overlapping simulation with learning on separate streams would gain little. Removing launch gaps gains more. Warp's own step graph cannot be launched inside a PyTorch capture, so we capture Warp's kernels directly into the PyTorch graph on a shared stream, and the constraint solver's iteration loop becomes a conditional graph node. Episode ends are handled with fixed shapes: random reset frames are drawn for every robot and written only where the episode ended, which removes data-dependent indexing and host synchronizations.

\paragraph{Check of the reset path}
We compare all simulator state fields after a masked reset with those after the original reset. They agree bit for bit except for joint velocities, which differ by at most $2.4\times10^{-7}$.

\paragraph{Time split of the release recipe}
Over three seeds, the 3.38 minutes consist of about 9\,s of startup and compilation, 167\,s to the first checkpoint at return 0.82 and 26\,s of tail.

\section{Tracking Accuracy}
\label{app:accuracy}
\input{tables/accuracy}

\section{Lineage of the Recipe}
\label{app:lineage}
\input{tables/lineage}
\input{tables/earlier}

\section{Negative Results}
\label{app:neg}
\begin{table}[t]
\centering\small
\caption{\textbf{Interventions that did not shorten training.} Iterations to 0.82 relative to the configuration they were added to, one seed unless stated.}
\label{tab:neg}
\setlength{\tabcolsep}{4pt}
\begin{tabular}{@{}p{5.6cm}p{4.3cm}p{5.8cm}@{}}
\toprule
Intervention & Effect & Reading \\
\midrule
Batch $\times2$, $\times4$ (teacher regression) & $-3\%$ updates & noise is not limiting \\
Prioritized WM windows; prioritized imagination starts & within seed noise & reweighting changes variance only \\
Planning (CEM) teacher + behavior cloning, 4 variants & no strict-target speed-up; 1.6--1.7$\times$ wall & $\kappa\approx1$; the student does not learn a margin \\
CEM data at failure points into the world model & cliff error $-21$ to $-48\%$; policy worse where taught & model accuracy at failures is not the limit \\
Critic slope supervision; critic variants & no gain; 2$\times$ slower per iteration & slope is not limiting \\
Mixed pathwise/likelihood-ratio estimator & offline $\hat\lambda=0.01$--$0.06$; zeroth-order term 100--1600$\times$ noisier & pathwise variance does not grow exponentially \\
$\lambda$-return actor; gradient decay; $H\in\{16,24\}$ & $+27\%$ to neutral & trades model error for critic error \\
No critic & $+50\%$, never reaches 0.84 & the value beyond $H$ is needed \\
Width $\times2$ & $-22\%$ iterations, $+27\%$ wall & cost grows faster than updates fall \\
Policy 1024-512-256 & $+26\%$ iterations, equal wall, more tail failures (2 seeds) & less capacity for the hardest motions \\
WM 5$\to$2 batches, critic 5$\to$2 epochs & $-25\%$ cost per iteration; 0.82 not reached by 2000 & proxy training sets signal quality \\
Lookahead ($k=10$, $\alpha=0.5$) & $+60\%$ & parameter averaging in the loop hurts \\
Cosine step decay to 4000 iterations & $+3\%$ & a fixed schedule ignores progress \\
Sharpness-weighted loss; Fourier features (teacher) & $+17$ to $+21\%$ & spectral bias is unchanged by reweighting \\
Training-condition failure resets $\times1.7$; sampler cap; reference floor repair & tail unchanged & wrong measurement condition (\cref{app:mismatch}) \\
\bottomrule
\end{tabular}
\end{table}

%% file: tables/accuracy.tex
\begin{table}[t]
\centering\small
\caption{\textbf{Tracking accuracy of the release recipe.} Three seeds, mean (range), each checkpoint rolled out four times under the evaluation protocol. Errors are averaged over all control steps of all 40 clips. Body position: RMS error of the 14 tracked bodies with the reference re-anchored to the robot root, as in the reward; root drift: global root position error, accumulated over clips of 3.7 minutes on average without any global position feedback; foot slip: horizontal foot speed in contact (the reference motion itself has 10.2\,cm/s); contact agreement: share of foot frames whose contact state matches the reference.}
\label{tab:accuracy}
\begin{tabular}{@{}lcc@{}}
\toprule
Metric & At the target & Iteration 3500 \\
\midrule
Body position (cm) & 4.10 (4.02--4.14) & 3.45 (3.40--3.49) \\
Body orientation (deg) & 8.1 (7.9--8.3) & 6.9 (6.8--7.1) \\
Joint angle (deg) & 4.0 (4.0--4.1) & 4.2 (4.1--4.4) \\
Root drift (cm) & 93 (89--97) & 75 (73--77) \\
Foot slip (cm/s) & 7.1 (7.0--7.2) & 6.7 (6.6--6.7) \\
Contact agreement & 0.747 (0.744--0.749) & 0.764 (0.763--0.765) \\
Return & 0.840 (0.837--0.843) & 0.861 (0.858--0.863) \\
\bottomrule
\end{tabular}
\end{table}

%% file: tables/lineage.tex
\begin{table}[t]\centering\small
\caption{\textbf{From InstantTrack v1.0 to \bm{}} (RTX 4090, $40\times4$). Iterations and minutes to a mean return of 0.82, cost per iteration, and the iteration of the strict target (three consecutive clean checkpoints; per seed). $^\dagger$500-iteration grid. $^\ddagger$plus probe sweeps. From row A on, checkpoints every 100 iterations and a synchronous probe (about 20\,ms) are included, so ms/it rises there although the graphs make each iteration cheaper (paired measurement in \cref{sec:cost}).}
\label{tab:lineage}
\begin{tabular}{@{}lccccc@{}}\toprule
Configuration & seeds & it.\ 0.82 & min & ms/it & strict it. \\ \midrule
InstantTrack v1.0 (our run) & 2+1 & 4997 & 15.3 & 195 & 6167 (19.6)$^\dagger$ \\
+ semi-implicit WM, mirror & 3+1 & 4066 & $\approx$12 & 190 & -- \\
+ Muon (policy, WM) & 2 & 2598 & 7.6 & 194 & -- \\
+ reference residual, no warm-up & 1 & 1732 & 5.8 & 195 & 2750 (9.1) \\
+ sampler/contact fixes (nb) & 3 & 1587 & 5.5 & 195 & 2750 / 2500 / --  \\
+ update graphs, allocation & 3 & 1549 & 4.9 & 172$^\ddagger$ & 2500 / 3000 / 3000 \\
+ batch 2048 & 3 & 1701 & 4.0 & 122$^\ddagger$ & 2500 / 2250 / 2750 \\
+ graph capture of the whole iteration (A) & 3 & 1742 & 4.0 & 142$^\ddagger$ & 2800 / 2900 / 2200 \\
+ step-size schedule (B) & 3 & 1407 & 3.2 & 141$^\ddagger$ & 2100 / 2500 / 1800 \\
\rowcolor{ptsrow}\textbf{+ probe fixes, cutoff at 1400 (\bm{})} & 3 & 1306 & 2.9 & 135$^\ddagger$ & 1500 / 1500 / 1600 \\
\bottomrule\end{tabular}\end{table}

%% file: tables/earlier.tex
\begin{table}[t]
\centering
\caption{\textbf{Earlier recipes on the RTX 4090} (same protocol as \cref{tab:main}). A: whole-iteration graphs, constant step size; B: A with the progress-triggered step size; C: B without allocation. The probe of A--C was synchronous, started at iteration 1250 and swept every 250 iterations. \emph{6 clean checkpoints}: a stricter variant of the target rule; ``--'': not reached within 3500 iterations.}
\label{tab:earlier}
\small
\setlength{\tabcolsep}{3.2pt}
\resizebox{\linewidth}{!}{%
\begin{tabular}{@{}l c c c c c c c@{}}
\toprule
Configuration & ms/it & Iterations to 0.82 & First $\ge$0.82 & 3 clean checkpoints & 6 clean checkpoints & Tail failures & Final return \\
\midrule
No allocation (C) & 121 & 1439 / 1302 / 1479 & 3.2 / 3.0 / 3.2 (3.1) & 6.4 / 6.2 / 5.6 (6.1) & -- / -- / 5.6 (--) & 23 / 22 / 10 & 0.862 \\
Constant step size (A) & 142 & 1778 / 1679 / 1768 & 4.2 / 3.9 / 4.3 (4.1) & 6.6 / 6.7 / 5.1 (6.1) & 6.6 / 6.7 / 5.1 (6.1) & 9 / 14 / 5 & 0.862 \\
Progress-triggered step (B) & 141 & 1344 / 1331 / 1547 & 3.2 / 3.2 / 3.7 (3.3) & 4.8 / 5.7 / 4.1 (4.9) & 4.8 / 5.7 / 4.1 (4.9) & 2 / 10 / 2 & 0.861 \\
\ours \textbf{BlinkMimic (release)} & 135 & 1292 / 1456 / 1169 & 2.89 / 3.28 / 2.64 (2.94) & 3.32 / 3.28 / 3.53 (3.38) & 5.36 / -- / 3.53 (--) & 2 / 6 / 0 & 0.861 \\
\bottomrule
\end{tabular}}
\end{table}

%% file: appendix/tests.tex
\FloatBarrier
\section{Full Experiment Record}
\label{app:record}
\input{tables/record}

%% file: tables/record.tex
\small
\begin{longtable}{@{}p{4.0cm}p{0.9cm}p{5.1cm}p{5.0cm}@{}}
\caption{\textbf{All learning experiments behind this paper.} Iterations to 0.82 on $40\times4$ unless stated; ``1e'' = single evaluation (about $\pm500$ iterations of noise). Configurations as in \cref{tab:lineage}. Seeds: number of training seeds.}\label{tab:record}\\
\toprule
Experiment & Seeds & Result & Reading \\
\midrule\endfirsthead
\toprule Experiment & Seeds & Result & Reading \\ \midrule\endhead
\multicolumn{4}{@{}l}{\emph{Fit phase: signal side}}\\
Semi-implicit WM (on v1.0) & 3 & $-13\%$ & smaller Jacobian error \\
+ mirror augmentation (smir) & 3+1 & $-18.6\%$ for both together (4066 vs.\ 4997) & symmetry prior; WM direction cosine 0.73$\to$0.80 \\
WM Muon (alone / on smir+policy Muon) & 1 / 2 & $-21.9\%$ / $-48\%$ total & WM loss at 500 halved \\
Reference residual (on smir+Muon) & 1 & $-22\%$ iterations, $-27\%$ updates & RL-only gain (teacher unchanged) \\
No WM-only warm-up & 1 & $-11\%$ iterations, $+4\%$ wall to 0.82 & kept for simplicity \\
Oracle (late) world model, warm-up kept & 1 & $-4\%$; strict unchanged & map quality not limiting \\
No critic & 1 & $+50\%$; 0.84 never reached & critic value needed \\
Critic: alive-only / bowl prior / slope supervision & 1 each (1e) & $+4\%$ / $-6\%$ / $-3\%$ at half speed & critic slope not limiting \\
$\lambda$-return actor; gradient decay 0.95 & 1 (1e) & $+27\%$; $-8\%$ with lower return & chain shortening trades errors \\
Imagination action noise 0.1 & 1 (1e) & $\pm0$ & -- \\
Mixed pathwise/LR estimator (offline) & 3 ckpts & $\hat\lambda=0.013$--0.064; ZO term 100--1600$\times$ noisier; one-step gain negative & no variance blow-up to fix \\
Critic state-gradient training for $H=8$ (offline) & 2 ckpts & $\partial V$ correlation $\le0.41$ (needed 0.5) & rejected \\
Planning teacher (CEM) + BC, 4 variants & 1 each & strict not faster; 1.6--1.7$\times$ wall & teacher passes, student lacks margin \\
CEM data into WM & 1 & cliff error $-11$ to $-55\%$; policy worse where taught & better map $\ne$ better policy \\
FK/contact inputs to WM (offline) & 2 & violent-moment error $-12$ to $-22\%$ (needed $-50\%$) & rejected \\
\multicolumn{4}{@{}l}{\emph{Fit phase: step side}}\\
Policy Muon (on v1.0) & 2 & $-36\%$; teacher regression 3494$\to$1478 & update geometry \\
SOAP vs.\ Muon (teacher regression) & 1 & 1505 vs.\ 1478 & preconditioning exhausted \\
Two updates per iteration (teacher regression) & 1 & $-38\%$ iterations, $+24\%$ updates & update-bound \\
Teacher regression, 5/10/20/40 clips & 1 & 998 / 1336 / 1427 / 1499 updates & time goes to shared skills \\
Width $\times2$ (teacher / RL) & 2 / 1 & $-27\%$ updates / $-22\%$ iterations, $+27\%$ wall & capacity helps per update, not per second \\
Width $\times0.5$ (teacher) & 1 & $+87\%$ & -- \\
Policy 1024-512-256 (RL) & 2 & $+26\%$ iterations, equal wall, tail failures 8/13 vs.\ 2/5 & capacity for hard motions \\
Residual-LN, trunk, Fourier features, sharpness weighting & 1 each & $\pm0$ to $+21\%$ & network form not limiting \\
Lookahead ($k{=}10$, $\alpha{=}0.5$) & 1 & $+59\%$ & -- \\
LR $\times2$ (teacher / RL, batch 2048) & 1 / 1 & $-12\%$ / $-25\%$ & step size has headroom \\
LR screen $\times2,3,4,6,8$ (750 it) & 1 each & return at 500: 0.606 / 0.639 / 0.659 / 0.630 / 0.603 & ceiling $\times4$ \\
Cosine LR to 4000 & 1 & $+3\%$, strict unchanged & fixed schedule \\
LR $\times4$, progress-triggered decay & 1 (+3 in \cref{tab:earlier}) & 1349 vs.\ 1833 ($-26\%$) & see \cref{sec:exp-why} \\
\multicolumn{4}{@{}l}{\emph{Batch and cost}}\\
Batch 2048 & 3 & $+6$/$+7$/$+16\%$ iterations, $-17\%$ wall & above the knee \\
Batch 1024 / 512 & 1 / 1 & $+21\%$ / $+45\%$ (512 predicted 1.44--1.66$\times$) & knee; more late failures \\
Policy gradient on 1024, critic on 4096 & 1 & $+17\%$ & knee from policy-gradient noise \\
WM 2 batches, critic 2 epochs & 1 & $-25\%$ cost, 0.82 not reached by 2000 & proxy training sets signal quality \\
Batch $\times2$/$\times4$ (teacher regression) & 1 & $-3\%$ & not noise-limited \\
\multicolumn{4}{@{}l}{\emph{Tail phase}}\\
Hardest clip trained alone & 1 & 16/16 failures at 750--1000; 0/128 at 1250--3000 & learnable \\
Forced 10\% reset share before its hard spot & 1 & 0/256 at 6 checkpoints vs.\ 19\% / 3.9\% / 2.3\% & tail = practice share \\
Reference floor repair; deployment conditions in 10\% of training; sampler cap; failure resets $\times1.7$ & 1 each & no improvement & \cref{app:mismatch} \\
Policy anchoring to EMA & 2 & drift $-78$ to $-84\%$, failures unchanged & drift not the cause \\
Deployment-condition allocation (vs.\ none) & 3 & see \cref{sec:exp-tail} & -- \\
Probe share 0.3 / 0.5 instead of 0.15 (release recipe) & 3 / 6 & 3-checkpoint rule $-8\%$ / $-6\%$; iterations to 0.82 $-10\%$ / $-10\%$ & within seed spread; not adopted \\
\multicolumn{4}{@{}l}{\emph{Release recipe}}\\
Leave-one-out ablation, 7 rows & 3 each & see \cref{tab:ablation} & -- \\
Step-size cutoff at 1000 / 1200 instead of 1400 & 3 / 3 & iterations to 0.82 $-3\%$ / $-6\%$ & within seed spread \\
Reference inverse-dynamics feedforward on the residual ($s=1$ / $0.5$) & 3 / 3 & iterations to 0.82 $-6\%$ / $-8\%$ & within seed spread; not adopted \\
Twice / four times the updates per simulated step & 1 / 1 & 3-checkpoint rule 2700 vs.\ 1500 iterations / 0.80 at 1800 with 4 clips failing & rejected \\
\bottomrule
\end{longtable}